\documentclass{article}

    \usepackage[preprint]{neurips_2025}

\usepackage[utf8]{inputenc} % allow utf-8 input
\usepackage[T1]{fontenc}    % use 8-bit T1 fonts
\usepackage{hyperref}       % hyperlinks
\usepackage{url}            % simple URL typesetting
\usepackage{booktabs}       % professional-quality tables
\usepackage{amsfonts}       % blackboard math symbols
\usepackage{nicefrac}       % compact symbols for 1/2, etc.
\usepackage{microtype}      % microtypography
\usepackage{xcolor}         % colors

\usepackage{tcolorbox}
\usepackage{enumitem}
\usepackage{algorithmic}
\usepackage{algorithm}
\usepackage{multirow}%
\usepackage{mathrsfs}%
\usepackage[title]{appendix}%
\usepackage{textcomp}%
\usepackage{manyfoot}%
\usepackage{listings}%
\usepackage{anyfontsize}
\usepackage{color}
\usepackage{bbding}
\usepackage{longtable}
\usepackage{supertabular}
\usepackage{colortbl}
\usepackage{tocloft}
\usepackage{wrapfig} 
\usepackage{mathrsfs}
\usepackage{pifont} 
\usepackage{animate}
\usepackage{amsmath}
\usepackage{amssymb}
\usepackage{mathtools}
\usepackage{amsthm}

\usepackage[capitalize,noabbrev]{cleveref}

\theoremstyle{plain}

\theoremstyle{definition}

\theoremstyle{remark}

\title{Weight Spectra Induced Efficient Model Adaptation}

\author{%
  Chongjie Si$^1$, Xuankun Yang$^1$, Muqing Liu$^2$, Yadao Wang$^3$ \\ \textbf{Xiaokang Yang$^1$, Wenbo Su$^3$, Bo Zheng$^3$, Wei Shen$^1$} \\
  $^1$Shanghai Jiao Tong University, $^2$Southeast University, $^3$Alibaba Group \\
  \texttt{\{chongjiesi, wei.shen\}@sjtu.edu.cn} \\
}

\begin{document}

\maketitle

\begin{abstract}

Large-scale foundation models have demonstrated remarkable versatility across a wide range of downstream tasks. 
However, fully fine-tuning these models incurs prohibitive computational costs, motivating the development of Parameter-Efficient Fine-Tuning (PEFT) methods such as LoRA, which introduces low-rank updates to pre-trained weights.
Despite their empirical success, the underlying mechanisms by which PEFT modifies model parameters remain underexplored.
In this work, we present a systematic investigation into the structural changes of weight matrices during fully fine-tuning. 
Through singular value decomposition (SVD), we reveal that fine-tuning predominantly amplifies the top singular values while leaving the remainder largely intact, suggesting that task-specific knowledge is injected into a low-dimensional subspace.
Furthermore, we find that the dominant singular vectors are reoriented in task-specific directions, whereas the non-dominant subspace remains stable. 
Building on these insights, we propose a novel method that leverages learnable rescaling of top singular directions, enabling precise modulation of the most influential components without disrupting the global structure. 
Our approach achieves consistent improvements over strong baselines across multiple tasks, highlighting the efficacy of structurally informed fine-tuning.

\end{abstract}

\section{Introduction}

The advent of foundation models \cite{brown2020language, kirillov2023segment,devlin2018bert,liu2019roberta} has showcased exceptional efficacy and versatility across artificial intelligence community. 
Traditionally, leveraging pre-trained models for specific applications involves fully fine-tuning all parameter \cite{ma2024segment, raffel2020exploring, qiu2020pre}.
Nonetheless, with the increasing complexity and number of parameters in these models, this traditional method of fully fine-tuning has become increasingly untenable, leading to significant resource demands.

To address this problem, recent years has witnessed a tremendous success in Parameter Efficient Fine-tuning (PEFT) \cite{zhang2022adaptive,si2024see,pfeiffer2020adapterfusion,houlsby2019parameter,hu2021lora,he2021towards,si2024flora}, which focuses on adjusting only a minimal fraction of the model's parameters while still achieving or surpassing the results of full parameter adjustments.
Among various PEFT methods, LoRA \cite{hu2021lora} has become increasingly favored for its adaptability. 
Specifically, for a frozen weight matrix $\mathbf{W}\in\mathbb{R}^{n\times m}$, LoRA learns an additional low-rank term $\Delta\mathbf{W} = \mathbf{A}\mathbf{B} \in\mathbb{R}^{n\times m}$, where $\mathbf{\mathbf{A}} \in \mathbb{R}^{n\times r}$ and $\mathbf{B}\in\mathbb{R}^{r\times m}$ are two low-rank matrices with $r\ll\{n,m\}$. 
This additional term is added to the frozen weight, with the form as 
\begin{equation} 
    \mathbf{W} \rightarrow \mathbf{W} + \Delta \mathbf{W},
\end{equation}
where $\mathbf{W}$ is the updated matrix. 
The matrix $\mathbf{A}$ is initialized with the uniform Kaiming distribution \cite{he2015delving}, whereas matrix $\mathbf{B}$ is initially set to zero.
Throughout the fine-tuning process, the matrices $\mathbf{A}$ and $\mathbf{B}$ are updated while $\mathbf{W}$ remains unchanged. 
Following LoRA, various methods have been introduced based on low-rank adaptation, to facilitate parameter efficient tuning through the application of low-rank decomposition \cite{si2024flora, feng2024trilora, zhang2022adaptive, wu2024mixture}.

Despite the empirical success of these methods, a deeper understanding of how pre-trained weights evolve during fine-tuning remains limited.
In this work, we conduct a systematic analysis of both the pre-trained and fine-tuned weight matrices to shed light on the internal mechanisms driving PEFT. 
Specifically, we examine the singular value decomposition (SVD) of the weight matrices and uncover striking structural regularities.
We find that the singular value spectra of the pre-trained and fine-tuned weights exhibit substantial overlap, with the primary distinction being that the top singular values of the fine-tuned weights are amplified, while the remaining singular values remain largely unchanged.

To further probe this phenomenon, we analyze the associated singular vectors (i.e., task-specific directions \cite{si2024unleashing}).
Interestingly, we observe that the top singular vectors across models are nearly orthogonal, indicating that fine-tuning introduces substantial alterations in these dominant directions, often unrelated to those in the pre-trained model. 
In contrast, the remaining singular vectors exhibit high mutual similarity, suggesting that these subspaces remain largely preserved during adaptation.
This contrast implies that new knowledge is primarily injected into a low-dimensional subspace, while the majority of the pre-trained structure is retained.

Building upon these observations and analyses, we propose a novel method that leverages the structural insights revealed through the singular value decomposition. 
Specifically, we posit that the low-rank updates employed by LoRA provide an effective mechanism for modulating the singular values of the underlying weight matrices. 
This formulation inherently aligns with the observation that task-specific knowledge is concentrated along the top singular directions \cite{meng2024pissa}.
Motivated by this, we further introduce a simple yet effective strategy: directly rescaling the top singular vectors of the pre-trained weights. 
By applying learnable scaling factors to these dominant directions, we enable the model to more precisely adjust the task-specific subspace without perturbing the broader representational structure. 
This targeted adjustment facilitates more efficient adaptation, as it focuses the capacity of LoRA-style updates on the most influential components of the model’s parameter space.
Extensive experiments have shown that our method can achieve superior performances to those of other SOTA methods.

\section{Related Work}
\subsection{Parameter Efficient Fine-Tuning}

The deployment of large-scale foundation models, often comprising billions of parameters, typically relies on full fine-tuning for adaptation to downstream tasks. 
However, this process incurs substantial computational and memory costs. 
To address this challenge, Parameter-Efficient Fine-Tuning (PEFT) methods have emerged as a promising alternative, aiming to preserve downstream performance while significantly reducing the number of trainable parameters and resource consumption \cite{zhang2022adaptive, si2024see, pfeiffer2020adapterfusion, houlsby2019parameter, hu2021lora, he2021towards, si2024flora, li2021prefix}.
Existing PEFT approaches can be broadly categorized into three paradigms.
(1) Adapter-based methods \cite{houlsby2019parameter, pfeiffer2020adapterfusion, he2021towards} insert lightweight, trainable modules into the layers of the transformer architecture. 
These modules are trained while keeping the backbone model fixed, allowing task-specific adaptation with minimal parameter updates.
(2) Prompt-based methods \cite{li2021prefix, shi2023dept, razdaibiedina2023residual} introduce learnable continuous vectors, either prepended to the input tokens (prompt tuning) or injected into the intermediate representations (prefix tuning), thereby steering the model behavior without modifying the backbone.
(3) Low-Rank Adaptation (LoRA) \cite{hu2021lora, zhang2022adaptive, si2024flora, si2024see} assumes that the weight updates required for downstream tasks lie in a low-dimensional subspace. 
LoRA decomposes the weight updates into low-rank matrices, enabling efficient task adaptation with negligible inference overhead and memory footprint. 
Unlike adapters or prompts, LoRA directly modifies the weight matrices in a low-rank form, thus facilitating more fine-grained control over the learned subspace.
Moreover, LoRA is highly flexible and can be seamlessly integrated with other PEFT techniques, such as adapters and prompt tuning, leading to further improvements in parameter efficiency and training scalability. This compositionality makes LoRA a particularly attractive design choice in modern PEFT frameworks.

\subsection{Metrics of Matrix Information Content}
Singular value-based metrics have been widely adopted to quantify the information content embedded in matrix representations, with prominent examples including effective rank \cite{roy2007effective} and spectral entropy \cite{cohen2021gradient}. 
In this work, we investigate the informational dynamics of weight matrices during fine-tuning using two complementary perspectives: (i) the distribution of singular values and (ii) the geometric transformation of the matrix space.

First, we analyze the singular value distributions derived from singular value decomposition (SVD), which captures the spectrum of energy concentration across principal components. 
This spectrum offers a proxy for the matrix’s representational capacity and structural complexity.
By monitoring the evolution of singular values before and after fine-tuning, we assess how information is redistributed across different dimensions of the weight matrix.
Second, to examine how the geometric structure of the parameter space evolves during fine-tuning, we compute the cosine similarity between the singular vectors of the pre-trained and fine-tuned weight matrices. 
This provides insight into the degree of alignment or reorientation in the learned subspaces, highlighting how fine-tuning modifies the directional flow of information in the model.
Together, these two analyses offer a comprehensive view of how fine-tuning alters both the magnitude (via singular values) and directionality (via singular vectors) of information in neural representations. 
The following section will provide a detailed empirical exploration of these phenomena.

\section{Spectral Analysis of Pre-trained and Fine-Tuned Weights}\label{sec observation}

In this section, we conduct a comprehensive analysis of the structural changes in model weights before and after fine-tuning. 
Specifically, we fine-tune the pre-trained LLaMA3-8B model \cite{llama3modelcard} on the Commonsense170K dataset \cite{hu2023llm} and examine how the weight matrices evolve across key components. 
Our study focuses on two complementary aspects: singular value distributions and the alignment of singular vector subspaces.

\begin{figure*}[!ht]
    \centering
    \includegraphics[width=\linewidth]{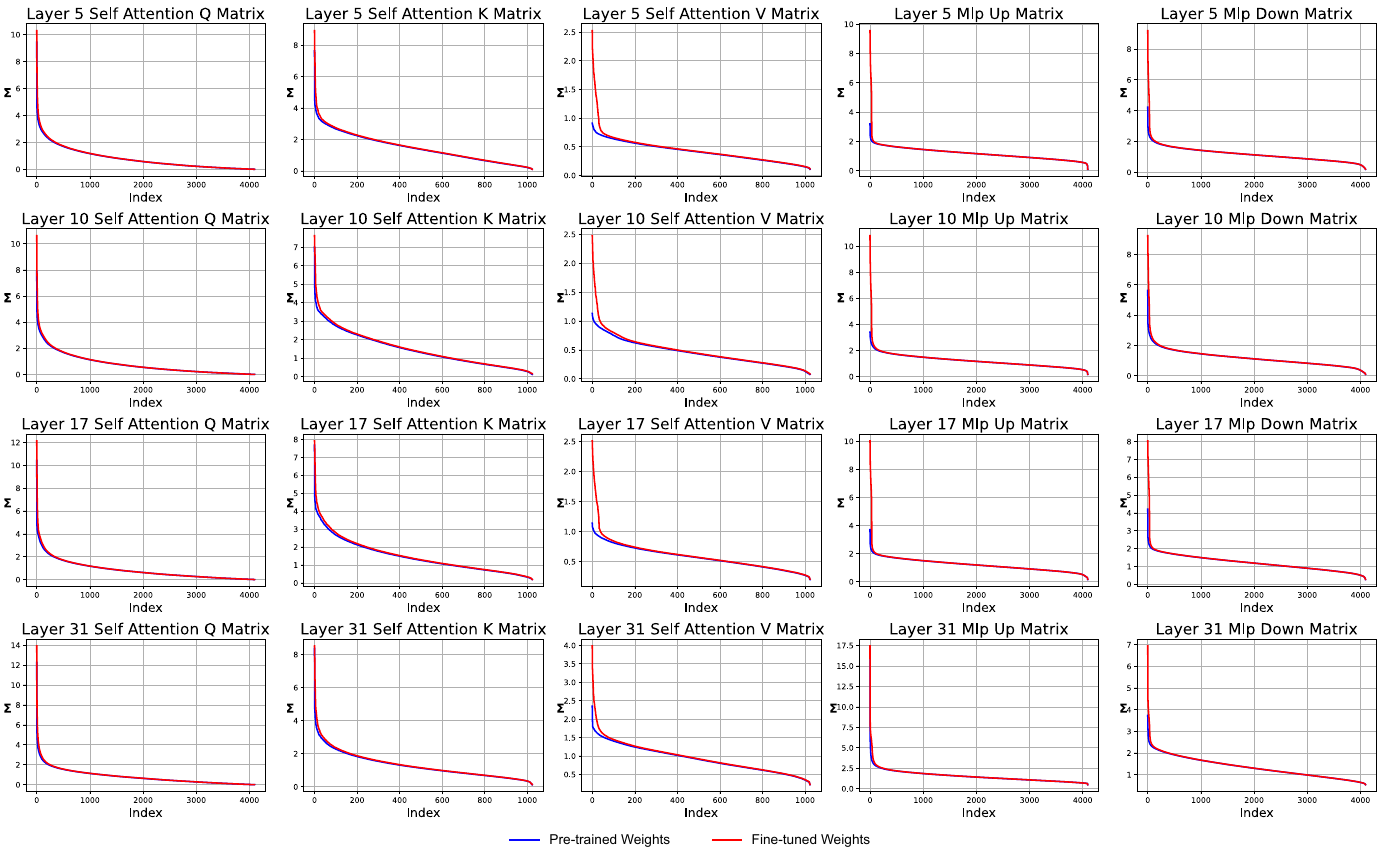}
    \caption{Singular value distributions of selected weight matrices before and after fine-tuning. We visualize the singular value spectra of attention Q, K, V matrices and MLP Up/Down projection matrices from randomly selected layers of LLaMA3-8B. Fine-tuning primarily amplifies the top singular values while leaving the rest largely unchanged.}
    \label{fig: singular value}
\end{figure*}

\subsection{Distributional Shifts in Singular Values}

Several widely used measures of matrix information content—such as effective rank and the entropy of the normalized singular value distribution—are fundamentally rooted in the singular values of the matrix. 
Motivated by this, we begin our exploration by analyzing the singular value distributions of the pre-trained and fine-tuned weight matrices.

We perform singular value decomposition (SVD) on selected weight matrices from randomly sampled layers.
We visualize the singular spectra of attention components (Q, K, V) and MLP projections (Up and Down) in Fig. \ref{fig: singular value}.
The results reveal consistent structural patterns. 
Across all examined modules, the singular value spectra of the pre-trained and fine-tuned weights show substantial overlap, suggesting that fine-tuning preserves the global spectral structure of the model. 
However, the top singular values in the fine-tuned weights are consistently amplified.
These dominant values correspond to the most task-relevant directions, indicating that fine-tuning reallocates representational emphasis without globally altering the rank or overall complexity of the matrix.

This selective amplification supports the core intuition behind parameter-efficient tuning strategies such as LoRA, where only a low-rank subspace is modified to encode task-specific knowledge while the majority of the model remains unchanged.

\subsection{Directional Shifts in Singular Vector Subspaces}

To complement our spectral analysis, we next investigate how the geometric structure of the weight matrices changes during fine-tuning. 
Specifically, we analyze the subspace similarity between the singular vectors of the pre-trained and fine-tuned weights to understand how the parameter space is reoriented during adaptation.

\begin{figure*}[!ht]
    \centering
    \includegraphics[width=\linewidth]{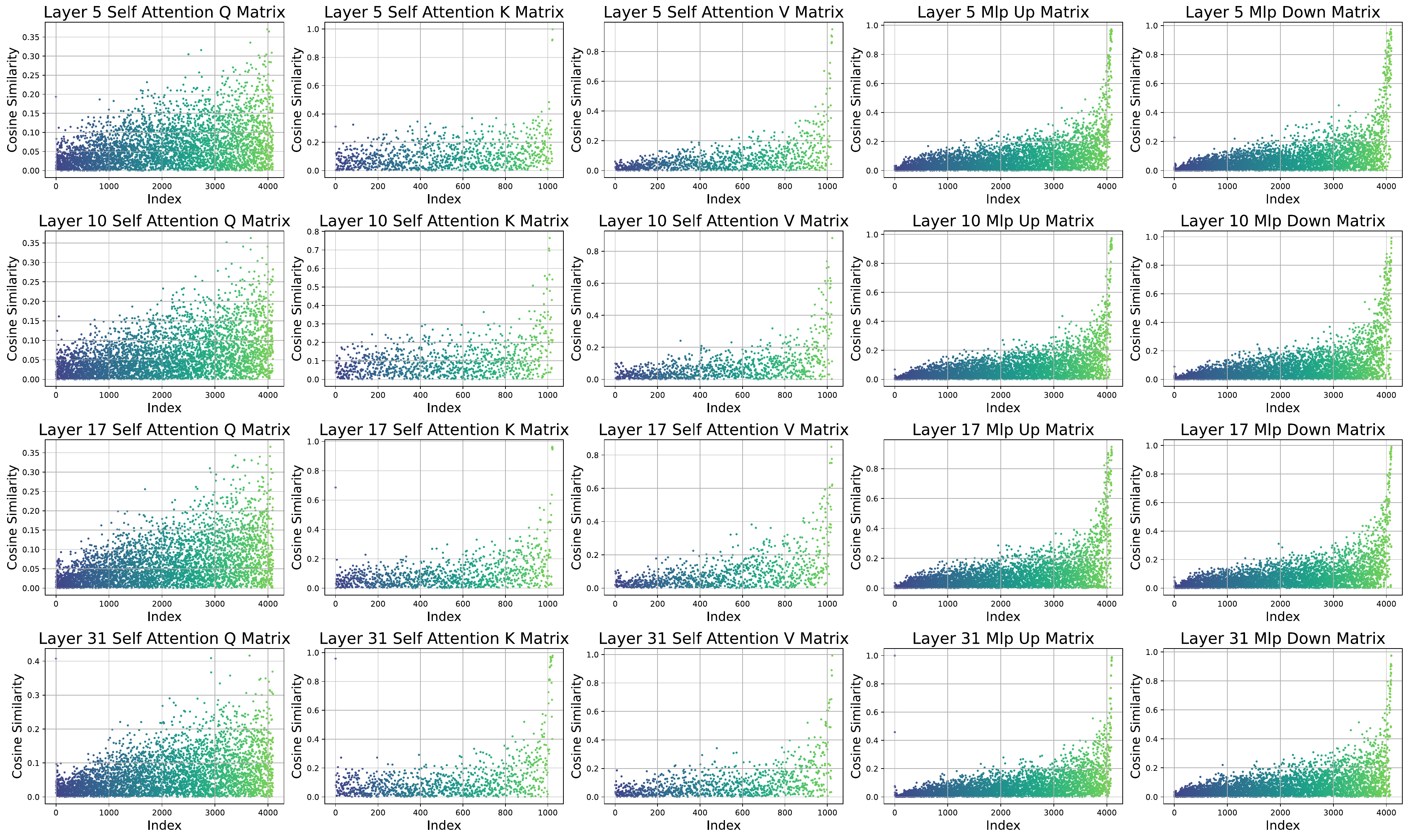}
    \caption{Cosine similarity between corresponding singular vectors of pre-trained and fine-tuned weights. For each selected layer and matrix, we compute the cosine similarity between singular vectors at the same index. Top singular directions exhibit low similarity, while lower directions remain closely aligned.}
    \label{fig: singular vector}
\end{figure*}

We compute the cosine similarity between corresponding singular vectors (i.e., same index) in the pre-trained and fine-tuned weight matrices across the same layers. 
This provides a fine-grained view of how each directional component is preserved or altered. 
The results are presented in Fig. \ref{fig: singular vector}.
We observe a striking divergence in similarity between top and bottom singular vectors.
The top singular directions—those associated with the largest singular values—tend to be nearly orthogonal, suggesting that fine-tuning induces substantial reorientation in the most important representational directions.
In contrast, the remaining singular vectors exhibit high similarity, indicating that much of the pre-trained geometry is retained.

These findings reinforce the hypothesis that fine-tuning primarily reshapes a compact, task-specific subspace while maintaining the broader structure of the pre-trained model. 
This observation aligns well with previous studies on intrinsic dimensionality in transfer learning \cite{aghajanyan2020intrinsic}.

\subsection{Spectral Stability and Subspace Reorientation}

Taken together, the analyses of singular values and singular vectors yield a unified perspective on how fine-tuning affects model parameters.
The preservation of the singular value spectrum, aside from amplification at the top, suggests that the information-carrying capacity of the matrices remains largely intact \cite{biderman2024lora}.
Simultaneously, the geometric misalignment in top singular vectors indicates the emergence of new, task-specific directions rather than mere scaling of existing ones \cite{meng2024pissa}.

This decoupling between value similarity and vector alignment suggests that fine-tuning introduces directional innovation without increasing global complexity. 
In other words, while the overall spectrum remains stable, the fine-tuned model reorients a small subset of directions to align with task-specific objectives.
This points to fine-tuning as a low-rank but geometrically transformative process-one that injects new knowledge through precise modifications to a limited number of structurally significant directions, while preserving the pre-trained scaffold elsewhere in the parameter space.

% This understanding serves as a direct motivation for our proposed method, which explicitly targets and modulates these top directions to enhance the effectiveness and interpretability of parameter-efficient fine-tuning.

\section{SpecLoRA: Enhancing Fine-Tuning via Principal Direction Modulation}

\subsection{Overview}

Inspired by the empirical observations presented in Sec. \ref{sec observation}, we propose a novel fine-tuning framework, termed Spectral-Directed LoRA (SpecLoRA). 
Our approach retains the foundational design of LoRA by applying a low-rank adaptation to frozen pre-trained weights, while explicitly incorporating a spectral perspective that enables the model to selectively modulate the dominant singular directions of the original parameter matrix.

Concretely, for a frozen pre-trained weight matrix $\mathbf{W} \in \mathbb{R}^{n \times m}$ ($n<m$ without loss of generalization), standard LoRA introduces a learnable low-rank residual $\Delta \mathbf{W} = \mathbf{A} \mathbf{B}$, where $\mathbf{A} \in \mathbb{R}^{n \times r}$ and $\mathbf{B} \in \mathbb{R}^{r \times m}$ with $r \ll \{n, m\}$.
The updated weight used for downstream inference is defined as:
\begin{equation}
\mathbf{W} \rightarrow \mathbf{W} + \Delta \mathbf{W} = \mathbf{W} + \mathbf{A} \mathbf{B}.
\end{equation}
This formulation can be interpreted as learning a low-rank subspace to encode task-specific information. 
However, as observed in our analysis, the top singular directions of the weight matrices undergo the most significant transformations during fine-tuning.
Moreover, these directions in the fine-tuned and pre-trained models are nearly orthogonal, suggesting that task-specific adaptation primarily occurs along a reoriented, low-dimensional spectral basis.

\subsection{Spectral-Directed Rescaling}

Motivated by these observations, we introduce a mechanism to explicitly adjust the top-$k$ singular directions of the pre-trained weight matrix. 
Our goal is to preserve the representational capacity of the pre-trained model while enabling targeted adaptation along the most task-relevant axes. 
Specifically, let the singular value decomposition of the frozen weight matrix $\mathbf{W}$ be:
\begin{equation}
\mathbf{W} = \mathbf{U} \boldsymbol{\Sigma} \mathbf{V}^\top,
\end{equation}
where $\mathbf{U} \in \mathbb{R}^{n \times n}$ and $\mathbf{V} \in \mathbb{R}^{m \times m}$ are orthogonal matrices of left and right singular vectors, and $\boldsymbol{\Sigma} = \operatorname{diag}(\sigma_1, \dots, \sigma_r, \dots)$ contains the singular values.
We denote $\mathbf{U}_{1:k} \in \mathbb{R}^{n \times k}$ as the matrix formed by the top-$k$ left singular vectors, corresponding to the largest $k$ singular values, and denote $\mathbf{U}_{k+1:n}$ the remained singular vectors. 
From linear algebra, we note that modifying a single coordinate of a nonzero vector is sufficient to alter its direction, provided the change is not colinear with the original vector. 
Hence, to introduce directional shifts while minimizing parameter overhead, we propose to modify only the top-$k$ rows of $\mathbf{U}_{1:k}$, i.e., the submatrix $\mathbf{U}_{1:k}^{(1:k)} \in \mathbb{R}^{k \times k}$.
We define a diagonal rescaling matrix $\mathbf{D} \in \mathbb{R}^{k \times k}$, and perform a structured modification:
\begin{equation}
\widetilde{\mathbf{U}}_{1:k}^{(1:k)} = \mathbf{D} \cdot \mathbf{U}_{1:k}^{(1:k)}.
\end{equation}
That is, each of the first $k$ rows of $\mathbf{U}_{1:k}$ is scaled individually along its own axis, effectively altering the orientation of the corresponding singular vectors. For the remaining rows ($i > k$), we retain the original components:
\begin{equation}
\widetilde{\mathbf{U}}_{1:k}^{(i)} = \mathbf{U}_{1:k}^{(i)}, \quad \forall i > k.
\end{equation}
This results in a modified matrix $\widetilde{\mathbf{U}}_{1:k} \in \mathbb{R}^{n \times k}$, where only the first $k$ rows have been altered.

\subsection{Final Formulation and Efficient Implementation}

Building upon the above formulation, we define the final fine-tuned weight matrix as a combination of two components:
\begin{equation}
\mathbf{W} \rightarrow\begin{bmatrix}
    \widetilde{\mathbf{U}}_{1:k} & \mathbf{U}_{k+1:n}
\end{bmatrix}\boldsymbol{\Sigma} \mathbf{V}^\top + \mathbf{A} \mathbf{B}.
\end{equation}
The first term provides a spectrally guided adaptation that explicitly adjusts the most influential singular directions, while the second term enables flexible task-specific learning in a complementary subspace.
However, explicitly computing the SVD and reconstructing the SVD components at each forward pass is computationally expensive.
To circumvent this bottleneck, we adopt a more efficient implementation that leverages a Hadamard product $\odot$ formulation:
\begin{equation}
\mathbf{W} \rightarrow (\boldsymbol{\Gamma} \odot \mathbf{W}) + \mathbf{A} \mathbf{B},
\end{equation}
where $\boldsymbol{\Gamma} \in \mathbb{R}^{n \times m}$ is a learnable spectral modulation mask defined as
\begin{equation}
\boldsymbol{\Gamma} =
\begin{bmatrix}
\underbrace{\begin{bmatrix}
\mathbf{d} & \mathbf{d} & \cdots & \mathbf{d}
\end{bmatrix}}_{k \text{ copies}} & \mathbf{1}_{k \times (m-k)} \\
\mathbf{1}_{(n-k) \times k} & \mathbf{1}_{(n-k) \times (m-k)}
\end{bmatrix},
\end{equation}
where $\mathbf{d}\in\mathbb{R}^k$ is a learnable scaling vector.
This implementation avoids direct SVD computation while still allowing the model to adjust the dominant directions of $\mathbf{W}$ in a fine-grained and learnable manner. 

\section{Experiment}

\subsection{Datasets and Models}
To validate the effectiveness of our method, we conduct comprehensive experiments on three representative tasks: natural language understanding, commonsense reasoning, and vision task.

For the natural language understanding (NLU) evaluation, we utilize the General Language Understanding Evaluation (GLUE) benchmark \cite{wang2018glue}, a widely adopted suite that covers a broad spectrum of language understanding tasks. The benchmark includes two single-sentence classification tasks, CoLA \cite{warstadt2019neural} and SST-2 \cite{socher2013recursive}, three similarity and paraphrase tasks, MRPC \cite{dolan2005automatically}, QQP \cite{wang2018glue}, and STS-B \cite{cer2017semeval}, and three natural language inference tasks, MNLI \cite{williams2017broad}, QNLI \cite{rajpurkar2016squad}, and RTE \cite{dagan2005pascal,bar2006second,giampiccolo2007third,bentivogli2009fifth}.
The details of these datasets are shown in Table. \ref{tab: glue dataset}.
We fine-tune DeBERTaV3-base \cite{he2021debertav3} on this task.

For commonsense reasoning task, we evaluate our method on a suite of eight sub-tasks, each associated with a dedicated benchmark dataset: BoolQ \cite{clark2019boolq}, PIQA \cite{bisk2020piqa}, Social IQA (SIQA) \cite{sap2019socialiqa}, HellaSwag \cite{zellers2019hellaswag}, WinoGrande \cite{sakaguchi2021winogrande}, ARC-e, ARC-c \cite{clark2018thinkarce}, and OpenBookQA (OBQA) \cite{mihaylov2018canobqa}. 
Following the experimental setup in \cite{hu2023llm}, we aggregate the training splits of all individual datasets into a unified training corpus, referred to as Commonsense170K. Model performance is then assessed separately on the test sets of each constituent task.
We fine-tune LLaMA3-8B \cite{llama3modelcard} for this task.

For vision task, we evaluate our method on VTAB-1k \cite{zhai2019visual}, a benchmark comprising 19 image classification tasks across three distinct categories: Natural, Specialized, and Structured. 
Each task provides 800 training samples and 200 validation samples, forming a total of 1,000 labeled examples per dataset.
Following the protocol established in prior works \cite{jia2022visual, jie2022convolutional, jie2023fact}, we fine-tune a pre-trained ViT-B/16 model \cite{dosovitskiy2020image} using the full set of 1,000 training and validation samples, and evaluate on the provided test set. Consistent with \cite{jia2022visual, lian2022scaling}, we adopt unnormalized image inputs, in line with the original VTAB implementation \cite{zhai2019visual}.

\subsection{Baselines and Implementation Details}

We compare our proposed method, SpecLoRA, against a range of state-of-the-art fine-tuning strategies, including: full fine-tuning, (IA)$^3$ \cite{liu2022few}, SSL and SSB \cite{si2024see}, BitFit \cite{zaken2021bitfit}, Series \cite{houlsby2019parameter}, Parallel \cite{pfeiffer2020adapterfusion}, AdaLoRA \cite{zhang2022adaptive},  LoRA \cite{hu2021lora}, DoRA \cite{liu2024dora}, AdaptFormer \cite{chen2022adaptformer}, NOAH \cite{zhang2022neural} and SSF \cite{lian2022scaling}.
Among adapter-based approaches, Series inserts trainable modules between the self-attention and feed-forward network (FFN) blocks, followed by residual connections. In contrast, Parallel adopts a more minimalistic architecture by placing adapters only after the FFN and LayerNorm components. 
For low-rank methods, following the protocol of \cite{zhang2022adaptive}, we apply LoRA, AdaLoRA, and DoRA uniformly across all learnable weight matrices.
Further implementation specifics can be found in the respective original works.
For SpecLoRA, we set $k=200$ for NLU and commonsense reasoning tasks, and $k=32$ for vision task. 
All experiments are conducted on NVIDIA H20 GPUs.

\subsection{Experiment Results}
Tables~\ref{tab: deberta results comparison}–\ref{tab: vtab} show the results of SpecLoRA across three benchmarks.
Across all settings, SpecLoRA demonstrates consistent and robust performance improvements over existing PEFT methods.

On the GLUE benchmark, SpecLoRA achieves the highest average score of 89.48, outperforming both LoRA and DoRA while updating only 0.18\% of the model parameters. 
In particular, SpecLoRA delivers notable gains on low-resource and structure-sensitive tasks such as CoLA (+1.91 over LoRA) and RTE (+2.88 over LoRA), highlighting its effectiveness in fine-tuning under constrained capacity by focusing on task-relevant spectral components.

On the commonsense reasoning benchmark, fine-tuned on LLaMA3-8B, SpecLoRA achieves the best overall accuracy of 85.5, surpassing DoRA and other strong baselines under the same parameter budget.
These results validate that spectral-aware adaptation enables better generalization across heterogeneous commonsense tasks.

On the VTAB-1K benchmark, SpecLoRA establishes a new state-of-the-art among PEFT methods with an average score of 76.7.
It outperforms strong visual adaptation baselines such as NOAH and SSF, while maintaining a comparable parameter footprint. 
SpecLoRA achieves strong performance across all three VTAB categories—Natural, Specialized, and Structured—demonstrating its general applicability and robustness across vision domains.

Taken together, these results consistently confirm the advantage of introducing spectral guidance into low-rank adaptation.
By selectively modulating the top singular directions, SpecLoRA achieves more effective task adaptation while preserving the representational integrity of the pre-trained model.

\begin{table*}[ht]
    \centering
    \renewcommand\arraystretch{0.95} 
    \caption {Results with DeBERTaV3-base fine-tuned on GLUE development set. ``FT'' represents fully fine-tuning.}
    \resizebox{\textwidth}{!}{
    \begin{tabular}{l | c| c c c c c c c c |>{\columncolor{gray!10}}c}
    % \hline\hline
    \toprule
         \multirow{2}{*}{\textbf{Method}} &  \multirow{2}{*}{\textbf{\% Params}} & \textbf{MNLI} & \textbf{SST-2} &\textbf{CoLA} & \textbf{QQP} & \textbf{QNLI} & \textbf{RTE} & \textbf{MRPC} & \textbf{STS-B} & \textbf{All}\\
         & & Acc & Acc & Mcc & Acc & Acc & Acc & Acc & Corr & Avg. \\ 
         \midrule
        
        FT & 100\% & 89.90 & 95.63 & 69.19 & 91.87 & 94.03 & 83.75 & 90.20 & 91.60 & 88.27 \\ \midrule
        
        (IA)$^3$ & 0.03\% & 89.44 & 95.52 & 67.01 & 89.01 & 91.80 & 79.42 & 88.23 & 90.79 & 86.40 \\  
         
         SSL & 0.02\% & 88.35 & 95.07 & 66.64 & 88.19 & 90.10 & 82.31 & 88.68 & 90.13 & 86.18 \\
        
         SSB & 0.05\% & 89.86 & 95.53 & 67.82 & 89.87 & 93.41 & 83.75 & 88.72 & 90.94 & 87.49 \\
         
        BitFit & 0.05\% & 89.37 & 94.84 & 66.96 & 88.41 & 92.24 & 78.80 & 87.75 & 91.35 & 86.21 \\ \midrule
        
         Series & 0.17\% & 90.10 & 95.41 & 67.65 & 91.19 & 93.52 & 83.39 & 89.25 & 91.31 & 87.73\\
        
        Parallel & 0.16\% & 89.89 & 94.72 & 69.06 & 91.05 & 93.87 & 84.48 & 89.71 & 91.38 & 88.02\\

         LoRA & 0.18\% & 90.03 & 93.92 & 69.15 & 90.61 & 93.37 &  87.01 & 90.19 & 90.75 & 88.13  \\  

        AdaLoRA & 0.18\% & 90.66 & 95.80 & 70.04 & 91.78 & 94.49 & 87.36 & 90.44 & 91.63 & 88.86 \\
        
        % FLoRA & 0.18\% & 90.60 & 96.00 & 70.20 & 91.40 & 94.46 & 88.09 & 90.93 & 91.96 & 89.21 \\
        
        DoRA & 0.22\% & 90.21 & 94.38 & 69.33 & 90.84 & 93.26 & 86.94 & 90.19 & 91.34 & 88.31 \\
        
         % LoRA-Dash & 0.18\% & 90.14 & 95.42 & 72.41 & 91.65 & 94.36 & 89.89 & 91.67 & 91.64 & {89.65} \\ 

         \rowcolor{gray!20}

        SpecLoRA & 0.18\% & 90.42 & 96.10 & 71.06 & 91.79 & 94.33 & 89.89 & 90.44 & 91.81 & 89.48 \\
         
        \bottomrule
    \end{tabular}
    }
    \label{tab: deberta results comparison}
\end{table*}

\begin{table*}[!ht]
 \renewcommand\arraystretch{1.0}
 \setlength{\tabcolsep}{2.4mm}
    \centering
    \caption{Results for LLaMA3-8B fine-tuned on commonsense reasoning tasks.}
    \resizebox{\textwidth}{!}{
    \begin{tabular}{ l c | c c c  c c c c c | l }
    \toprule
     \textbf{Method} & \textbf{Params(\%)} & \textbf{BoolQ} & \textbf{PIQA} & \textbf{SIQA} & \textbf{HellaS.} & \textbf{WinoG.} & \textbf{ARC-e} & \textbf{ARC-c} & \textbf{OBQA} & \multicolumn{1}{c}{\textbf{Avg.}} \\
    \toprule
    
    LoRA$_{r=16}$ & 0.35\% & 72.3 & 86.7 & 79.3 & 93.5 & 84.8 & 87.7 & 75.7 & 82.8 & 82.8 \\

    PISSA \cite{meng2024pissa} & 0.70\% & 67.1 & 81.1 & 77.2 & 83.6 & 78.9 & 77.7 & 63.2 & 74.6 & 75.4\\

    MiLoRA \cite{wang2024milora} & 0.70\% & 68.8 & 86.7 & 77.2 & 92.9 & 85.6 & 86.8 & 75.5 & 81.8 & 81.9 \\

    AdaLoRA & 0.35\% & 75.1 & 86.4 & 76.7 & 75.4 & 83.3 & 90.4 & 79.1 & 85.0 & 81.4 \\

    DoRA & 0.35\% & 74.5 & 88.8 & 80.3 & 95.5 & 84.7 & 90.1 & 79.1 & 87.2 & 85.0 \\

    \rowcolor{gray!20}

    SpecLoRA & 0.35\% & 74.6 & 89.8 & 80.9 & 95.5 & 85.3 & 90.1 & 80.3 & 87.2 & 85.5  \\
    
    \bottomrule
    \end{tabular}}
    \label{tab:results of commonsense}
\end{table*}

\begin{table}[ht]
  \centering
  \caption{Results on VTAB-1K benchmark.}
  \renewcommand\arraystretch{1} 
  \setlength{\tabcolsep}{0.6mm}
  \resizebox{\textwidth}{!}{%
    \begin{tabular}{c c | c c c c c c c | c c c c | c c c c c c c c | c }
      \toprule
      & & \multicolumn{7}{c}{\textbf{Natural}} & \multicolumn{4}{c}{\textbf{Specialized}} & \multicolumn{7}{c}{\textbf{Structured}} & \\
      
      & \rotatebox{90}{\# \textbf{Param (M)}} 
        & \rotatebox{90}{\textbf{Cifar100}} 
        & \rotatebox{90}{\textbf{Caltech101}} 
        & \rotatebox{90}{\textbf{DTD}} 
        & \rotatebox{90}{\textbf{Flower102}} 
        & \rotatebox{90}{\textbf{Pets}} 
        & \rotatebox{90}{\textbf{SVHN}} 
        & \rotatebox{90}{\textbf{Sun397}} 
        & \rotatebox{90}{\textbf{Camelyon}} 
        & \rotatebox{90}{\textbf{EuroSAT}} 
        & \rotatebox{90}{\textbf{Resisc45}} 
        & \rotatebox{90}{\textbf{Retinopathy}} 
        & \rotatebox{90}{\textbf{Clevr-Count}} 
        & \rotatebox{90}{\textbf{Clevr-Dist}} 
        & \rotatebox{90}{\textbf{DMLab}} 
        & \rotatebox{90}{\textbf{KITTI-Dist}} 
        & \rotatebox{90}{\textbf{dSpr-Loc}} 
        & \rotatebox{90}{\textbf{dSpr-Ori}} 
        & \rotatebox{90}{\textbf{sNORB-Azim}} 
        & \rotatebox{90}{\textbf{sNORB-Ele}} 
        & \rotatebox{90}{\textbf{Average}} \\
      
      \midrule
      
      \multicolumn{22}{l}{\textrm{\emph{Fully Fine-Tuning}}} \\
      \midrule
      
      Full  & 85.8 & 68.9 & 87.7 & 64.3 & 97.2 & 86.9 & 87.4 & 38.8 & 79.7 & 95.7 & 84.2 & 73.9 & 56.3 & 58.6 & 41.7 & 65.5 & 57.5 & 46.7 & 25.7 & 29.1 & 68.9 \\
      Linear & 0.04 & 64.4 & 85.0 & 63.2 & 97.0 & 86.3 & 36.6 & 51.0 & 78.5 & 87.5 & 68.5 & 74.0 & 34.3 & 30.6 & 33.2 & 55.4 & 12.5 & 20.0 & 9.6 & 19.2 & 57.6 \\
      
      \midrule
      
      \multicolumn{22}{l}{\emph{PEFT methods}} \\
      \midrule
      
      LoRA & 0.29 & 67.1 & 91.4 & 69.4 & 98.8 & 90.4 & 85.3 & 54.0 & 84.9 & 95.3 & 84.4 & 73.6 & 82.9 & 69.2 & 49.8 & 78.5 & 75.7 & 47.1 & 31.0 & 44.0 & 74.5 \\
      
      AdaptFormer & 0.16 & 70.8 & 91.2 & 70.5 & 99.1 & 90.9 & 86.6 & 54.8 & 83.0 & 95.8 & 84.4 & 76.3 & 81.9 & 64.3 & 49.3 & 80.3 & 76.3 & 45.7 & 31.7 & 41.1 & 74.7 \\  
      
      NOAH & 0.36 & 69.6 & 92.7 & 70.2 & 99.1 & 90.4 & 86.1 & 53.7 & 84.4 & 95.4 & 83.9 & 75.8 & 82.8 & 68.9 & 49.9 & 81.7 & 81.8 & 48.3 & 32.8 & 44.2 & 75.5 \\
      
      SSF & 0.20 & 69.0 & 92.6 & 75.1 & 99.4 & 91.8 & 90.2 & 52.9 & 87.4 & 95.9 & 87.4 & 75.5 & 75.9 & 62.3 & 53.3 & 80.6 & 77.3 & 54.9 & 29.5 & 37.9 & 75.7 \\
      
      \rowcolor{gray!10} 
      SpecLoRA & 0.30 & 72.5 & 92.1 & 71.6 & 99.1 & 91.0 & 89.4 & 55.8 & 87.5 & 95.4 & 83.9 & 74.7 & 83.4 & 64.5 & 52.5 & 81.9 & 86.1 & 53.4 & 37.8 & 44.4 & 76.7 \\
      
      \bottomrule
    \end{tabular}%
  }
  \label{tab: vtab}
\end{table}

\subsection{Ablation Study}

In the ablation study, we investigate several key factors that may influence the effectiveness of our method.
Specifically, we examine: (1) the impact of the number of selected singular vectors; (2) the relationship between the number of trainable parameters and downstream performance; and (3) whether modifying the top or bottom singular directions leads to better adaptation.
These analyses provide further insight into the design choices underlying SpecLoRA.

\subsubsection{Impact of the Number of Selected Top Singular Vectors}

\begin{figure}[ht]
    \centering
    \includegraphics[width=\linewidth]{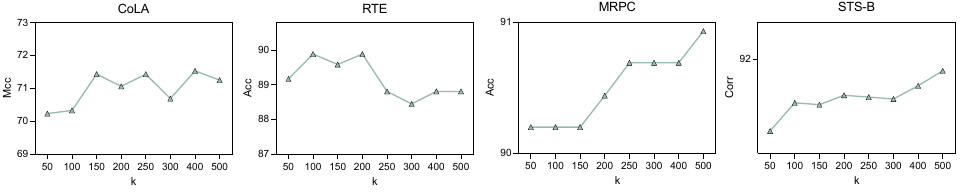}
    \caption{Ablation study on the number of trainable parameters (i.e., rank setting) of SpecLoRA.}
    \label{fig:k}
\end{figure}

Fig. \ref{fig:k} presents an ablation study on the number of selected top singular vectors $k$ used in SpecLoRA with $r=2$. 
Overall, we observe that model performance is relatively stable across a wide range of $k$ values, indicating the robustness of SpecLoRA to this hyperparameter.
On tasks such as MRPC and STS-B, performance improves steadily with larger $k$, suggesting that incorporating more top directions helps capture finer-grained semantics.
In contrast, on CoLA and RTE, performance peaks around $k=150$-$200$ and slightly fluctuates afterward, showing that moderate values of $k$ are sufficient to achieve strong results.
These results highlight that SpecLoRA is robust to the precise choice of $k$, and that a relatively small number of top directions already captures most task-relevant information.

\subsubsection{Impact of the Number of Trainable Parameters}

\begin{figure}[ht]
    \centering
    \includegraphics[width=\linewidth]{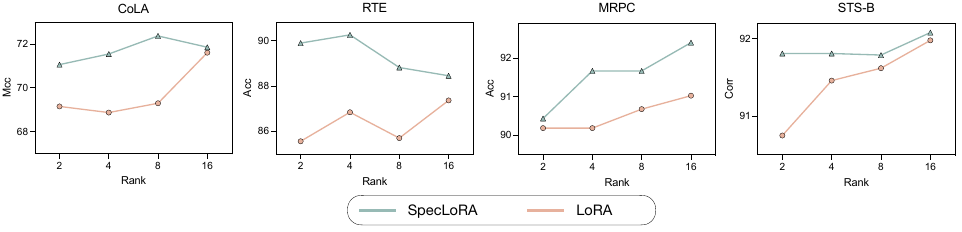}
    \caption{Ablation study on the number of trainable parameters (i.e., rank setting) of SpecLoRA.}
    \label{fig:rank}
\end{figure}

To assess the parameter efficiency of our method under different capacity budgets, we investigate the impact of the LoRA rank hyperparameter ($r = 2, 4, 8, 16$) on four representative GLUE tasks: CoLA, RTE, MRPC, and STS-B.
The results are shown in Fig. \ref{fig:rank}.
On MRPC and STS-B, performance improves steadily with rank, and SpecLoRA maintains a clear lead throughout.
Notably, on RTE, SpecLoRA achieves strong results even with a low rank, demonstrating better adaptation in low-resource scenarios. 
Overall, SpecLoRA consistently outperforms LoRA across all ranks and tasks, confirming its superior parameter efficiency.

\subsubsection{Impact of Top or Bottom Singular Directions}

To further understand the importance of spectral structure, we conduct an ablation study by comparing SpecLoRA, which modifies the top singular directions, with a variant that applies the same mechanism to the bottom singular directions. Results are summarized in Table \ref{tab: ablation top bottom}.
We observe that both spectral variants (Top and Bottom) outperform the standard LoRA baseline, indicating the general benefit of direction-aware adaptation. 
However, SpecLoRA, which operates on top singular directions, outperforms the bottom-direction variant and LoRA.
These results support our core hypothesis: the top singular directions capture task-relevant representational capacity, and adjusting them directly yields more effective and expressive adaptation under limited parameter budgets.

\begin{table*}[ht]
    \centering
    \renewcommand\arraystretch{0.95} 
    \caption {Ablation study on the location of directions on GLUE development set.}
    \resizebox{\textwidth}{!}{
    \begin{tabular}{l | c| c c c c c c c c |>{\columncolor{gray!10}}c}
    % \hline\hline
    \toprule
         \multirow{2}{*}{\textbf{Method}} &  \multirow{2}{*}{\textbf{\% Params}} & \textbf{MNLI} & \textbf{SST-2} &\textbf{CoLA} & \textbf{QQP} & \textbf{QNLI} & \textbf{RTE} & \textbf{MRPC} & \textbf{STS-B} & \textbf{All}\\
         & & Acc & Acc & Mcc & Acc & Acc & Acc & Acc & Corr & Avg. \\ 
         \midrule
        
         LoRA & 0.18\% & 90.03 & 93.92 & 69.15 & 90.61 & 93.37 &  87.01 & 90.19 & 90.75 & 88.13  \\  

        Bottom & 0.18\% & 90.14 & 95.99 & 70.60 & 91.83 & 94.29 & 88.81 & 89.46 & 91.63 & 89.09 \\

        SpecLoRA & 0.18\% & 90.42 & 96.10 & 71.06 & 91.79 & 94.33 & 89.89 & 90.44 & 91.81 & 89.48 \\
         
        \bottomrule
    \end{tabular}
    }
    \label{tab: ablation top bottom}
\end{table*}

\section{Limitation}
While SpecLoRA demonstrates strong empirical performance and is grounded in a principled spectral analysis, it is not without limitations.
Our approach currently assumes a fixed $k$ for all layers and weight matrices. 
While this simplifies implementation and parameter control, it may not be optimal across diverse network depths or parameter types. 
Adaptive or learned selection of $k$ per layer could further enhance flexibility and performance.

\section{Conclusion}

In this work, we present a principled study of parameter-efficient fine-tuning from a spectral perspective. 
Through a systematic SVD analysis of both pre-trained and fine-tuned weight matrices, we uncover that fine-tuning primarily amplifies the top singular values while preserving the remaining spectrum. 
Furthermore, we observe that the dominant singular vectors tend to reorient in task-specific directions, whereas the subdominant directions remain largely intact. 
Building on these insights, we propose SpecLoRA, which introduces learnable scaling on the top singular directions of the pre-trained weights, allowing the model to precisely and efficiently modulate task-relevant components without disturbing the overall representational space. 
This design enhances the expressivity and efficiency of adaptation while maintaining compatibility with existing PEFT pipelines.
Extensive empirical results across multiple benchmarks demonstrate that SpecLoRA consistently outperforms existing state-of-the-art PEFT methods.

\bibliography{main}
\bibliographystyle{plain}

%%%%%%%%%%%%%%%%%%%%%%%%%%%%%%%%%%%%%%%%%%%%%%%%%%%%%%%%%%%%

\newpage

\newpage
\appendix

\section*{Appendix}

We here present some experimental details.

\begin{table}[ht]
    \centering
    \caption{Hyper-parameter configurations for commonsense reasoning task.}
    \begin{tabular}{l c c c c}
        \toprule
        {Hyper-parameter} & {LoRA} & AdaLoRA & {DoRA} & SpecLoRA \\
        \midrule
        {Rank r} & \multicolumn{4}{c}{16} \\
        $\boldsymbol{\alpha}$ & \multicolumn{4}{c}{32} \\
        {Dropout} & \multicolumn{4}{c}{0.05} \\
        {Batch size} & \multicolumn{4}{c}{16} \\
        {Epochs} & \multicolumn{4}{c}{3} \\
        {Learning rate} & \multicolumn{4}{c}{3e-4}\\
        {Target module} & \multicolumn{4}{c}{\textit{q, k, v, up, down}} \\
        \bottomrule
    \end{tabular}
\label{tab: cr detail}
\end{table}

\begin{table*}[!ht]
    \centering
    \caption{Details of GLUE dataset.}
    \renewcommand\arraystretch{0.9}
    
    \resizebox{\textwidth}{!}{
    \begin{tabular}{l | l  c  c  c  c  c}
    \toprule
         Dataset & Task & \# Train & \# Dev & \# Test & \# Label & Metrics \\ \midrule
         \multicolumn{7}{c}{Single-Sentence Classification} \\ \hline
         
         CoLA & Acceptability & 8.5k & 1k & 1k & 2 & Matthews corr \\ \midrule
         
         SST-2 & Sentiment & 67k & 872 & 1.8k & 2 & Accuracy \\ \midrule
         
         \multicolumn{7}{c}{Similarity and Paraphrase} \\ \midrule

         MRPC & Paraphrase & 3.7k & 408 & 1.7k & 2 & Accuracy / F1 \\ \midrule

         QQP & Paraphrase & 364k & 40k & 391k & 2 & Accuracy / F1 \\ \midrule
         
         STS-B & Similarity & 7k & 1.5k & 1.4k & 1 & Pearson/ Spearman Corr \\  \midrule

        \multicolumn{7}{c}{Natural Language Inference} \\ \midrule
          
         MNLI & NLI & 393k & 20k & 20k & 3 & Accuracy \\ \midrule
         
         QNLI & QA/NLI & 108k & 5.7k & 5.7k & 2 & Accuracy \\ \midrule

         RTE & NLI & 2.5k & 276 & 3k & 2 & Accuracy \\
        
         \bottomrule
    \end{tabular}}
    \label{tab: glue dataset}
\end{table*}

\begin{table}
\renewcommand\arraystretch{0.9}
\centering
\caption{Hyper-parameter settings on NLU task.}
\resizebox{\textwidth}{!}{
\begin{tabular}{c | c c c c c c c c} 

\toprule

Hyper-parameter & MNLI & SST-2 & CoLA & QQP & QNLI & RTE & MRPC & STS-B\\ 

\midrule

Optimizer & \multicolumn{8}{c}{AdamW} \\ \midrule

Warmup Ratio & \multicolumn{8}{c}{0.1} \\ \midrule

LR schedule & \multicolumn{8}{c}{Linear} \\  \midrule

Rank $r$ & \multicolumn{8}{c}{2}\\ \midrule

LoRA alpha & \multicolumn{8}{c}{4} \\ \midrule

Max Seq. Len. & 256 & 128 & 64 & 320 & 512 & 320 & 320 & 128 \\ \midrule

Batch Size & 32 & 32 & 32 & 32 & 16 & 32 & 32 & 32 \\ \midrule

Learning Rate & 8e-4 & 4e-4 & 1e-3 & 5e-4 & 5e-4 & 1.2e-3 & 1e-4 & 1.8e-3 \\ \midrule

Epochs & 7 & 24 & 25 & 5 & 5 & 50 & 30 & 25  \\ 

\bottomrule

\end{tabular}}
\label{tab: nlu detail}
\end{table}

%%%%%%%%%%%%%%%%%%%%%%%%%%%%%%%%%%%%%%%%%%%%%%%%%%%%%%%%%%%%

\end{document}